\documentclass[journal]{IEEEtran}
\ifCLASSINFOpdf
\else
\fi

\ifCLASSINFOpdf
\usepackage[pdftex]{graphicx}
\graphicspath{{./fig/I&IV-prior/}{./fig/IV-match/}
			  {./fig/IV-reweighted/}{./fig/V-inpainting/}{./fig/V-multiscale/}
			  {./fig/V-denoise/}{./fig/multiscale/}{./fig/V-pure/}{./fig/V-synthesis/}
			  {./fig/V-parameters/}{./fig/V-color/}{./fig/V-failure/}}
\DeclareGraphicsExtensions{.pdf,.jpeg,.png,.jpg}
\else
\fi

\usepackage{times}
\usepackage{epsfig}
\usepackage{epstopdf}
\usepackage{graphicx}
\usepackage{graphics}
\usepackage[tbtags,fleqn]{amsmath}
\usepackage{amssymb}
\usepackage{array}
\usepackage{multirow}
\usepackage{color}
\usepackage{subfigure}
\usepackage{subfig}
\usepackage{url}
\usepackage{cite}
\usepackage{mathrsfs}
\usepackage{algorithm,algpseudocode}
\usepackage{amsfonts,amsthm}
\usepackage{bm}
\usepackage{mdwlist}
\usepackage{enumerate}
\usepackage{enumitem}
\DeclareMathOperator*{\argmin}{argmin}

\newcommand{\eg}{{\it e.g.}}
\newcommand{\ie}{{\it i.e.}}
\newcommand{\etal}{{\it et al.}}

\newcommand{\st}{\text{s.t. }}
\newcommand{\vect}[1]{\bm{#1}}
\newcommand{\set}[1]{\mathbb{#1}}
\newcommand{\matx}[1]{\bm{#1}}
\newcommand{\opt}[1]{\mathcal{#1}}

\begin{document}
%
\title{Image Cartoon-Texture Decomposition Using Isotropic Patch Recurrence}
%
%
%

\author{
Ruotao Xu, Yuhui Quan*, Yong Xu\thanks{All the authors are with the School of Computer Science and Engineering at South China University of Technology, China. Email: xu.ruotao@mail.scut.edu.cn, csyhquan@scut.edu.cn, yxu@scut.edu.cn. Asterisk indicates the corresponding author.}
}

%
%

\markboth{XXXXX}%
{}
%



\maketitle

\begin{abstract}
Aiming at separating the cartoon and texture layers from an image, cartoon-texture decomposition approaches resort to  image priors to model cartoon and texture respectively. In recent years, patch recurrence has emerged as a powerful prior for image recovery.
However, the existing strategies of using patch recurrence are ineffective to cartoon-texture decomposition, as both cartoon contours and texture patterns exhibit strong patch recurrence in images. To address this issue, we introduce the isotropy prior of patch recurrence, that the spatial configuration of similar patches in texture exhibits the isotropic structure which is different from that in cartoon,  to model the texture component. Based on the isotropic patch recurrence, we construct a nonlocal sparsification system which can effectively distinguish well-patterned features from contour edges. Incorporating the constructed nonlocal system into morphology component analysis, we develop an effective method to both noiseless and noisy cartoon-texture decomposition. The experimental results have demonstrated the superior performance of the proposed method to the existing ones, as well as the effectiveness of the isotropic patch recurrence prior.
\end{abstract}


%
\IEEEpeerreviewmaketitle

\section{INTRODUCTION}

A real image is usually the superposition of a cartoon component and a texture component. The cartoon component refers to the piecewise-constant geometrical parts of an image, including homogeneous regions, contours,  and sharp edges.
In contrast, the texture component is about the oscillating patterns of an image, such as fine structures and local repeating features. Cartoon-texture image decomposition is to separate the cartoon part and the texture part from an image, which plays an important role in computer vision, with a wide range of applications to image
restoration~\cite{bertalmio2003simultaneous,gilboa2006variational,ng2013coupled}, motion analysis~\cite{wedel2009improved}, image segmentation~\cite{cao2014segmentation}, image compression~\cite{hu2011multiple}, stereo matching~\cite{calderero2013recovering}, image editing~\cite{han2017cartoon,liang2018hybrid}, pattern recognition~\cite{figueiredo2015automated},
biomedical engineering~\cite{figueiredo2015automated},
remote sensing~\cite{aujol2005image,gilles2010properties,yanovsky2015separation}, 
etc.

The importance of cartoon-texture decomposition originates from the fact that the cartoon component and texture component exhibit significantly distinct characteristics and they often involve different operations in the processing, analysis and recognition of images. For instance, the texture component is preferred in optical flow estimation as it is often free of shading reflections and shadows~\cite{wedel2009improved}, while the cartoon part is often extracted for enhancing the stability of depth estimation~\cite{calderero2013recovering}.
In many image processing tasks (\eg~\cite{bertalmio2003simultaneous}), the cartoon regions and texture regions need separate treatments to guarantee the visual quality of results. Another example is object recognition, in which contour cues and texture features are extracted by different approaches (\eg~\cite{fadili2010image}).

In general, cartoon-texture decomposition requires solving the underdetermined system
\begin{equation}
\vect{f}=\vect{u}+\vect{v},
\end{equation}
where $\vect{f}\in\set{R}^{N}$ is the given image, $\vect{u}\in\set{R}^{N}$ is the cartoon component, and $\vect{v}\in\set{R}^{N}$ is the texture component.
Since the unknowns are much more than the equations, effective priors on both the cartoon part and texture part are needed to solve the system. In the existing approaches, the piecewise constant prior implemented by the total variation (TV) minimization is dominant in modeling cartoon parts (\eg~\cite{meyer2001oscillating,vese2003modeling,osher2003image,aujol2005image,buades2010fast,gilles2012multiscale,yin2005image,yin2005total,zhang2016convolutional,buades2016directional,papyan2017convolutional}) due to its advantages of inducing piecewise-constant functions. Regarding the texture component modeling, the low rank prior on texture patches is often used  (\eg~\cite{schaeffer2013low,fan2017cartoon}), and it has been recently shown in~\cite{ma2016group} that using the recurrence of local image patches, which is one powerful prior in image recovery (\eg~\cite{dabov2006image,dong2013nonlocally,quan2015data}), can lead to further improvement.

\begin{figure}[htbp!]
	\centering
	\includegraphics[width=\linewidth]{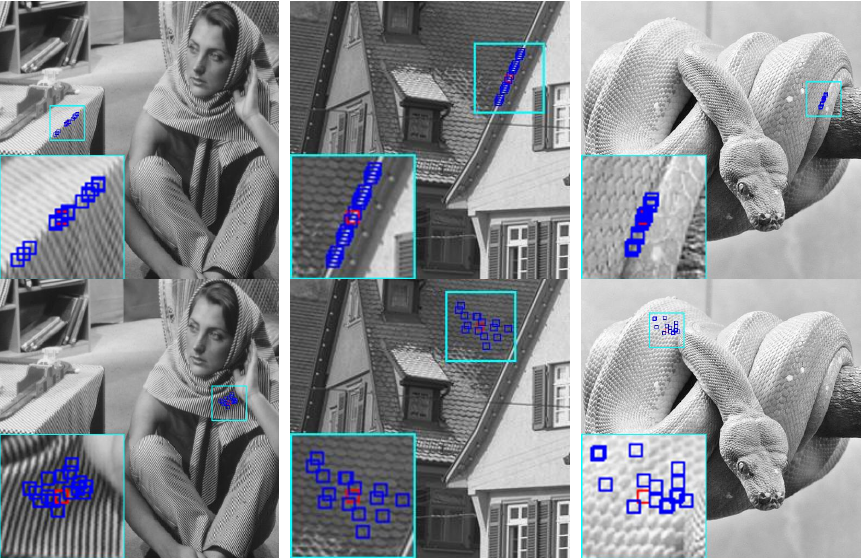}
	\caption{Anisotropic patch recurrence in cartoon (upper row) and isotropic patch recurrence in texture (bottom row).}
	\label{fig:prior_isotropy_anisotropy}
\end{figure}

However, the patch recurrence prior may also hold for cartoon regions. For instance, similar patches can also be found along a straight contour edge. In other words, there is some ambiguity when using patch recurrence to distinguish texture from cartoon. Thus, a direct call of the patch-recurrence-based strategies from image recovery may noticeably decrease the accuracy of the decomposition.
To further exploit the patch recurrence prior for image decomposition, in this paper, we introduce an additional prior on patch recurrence for better discriminating the cartoon component and texture component. The prior comes from the observation that, the neighboring similar patches around cartoon contours tend to align along a major direction, while the ones in texture regions are likely to scatter around. See Fig.~\ref{fig:prior_isotropy_anisotropy} for an illustration, in which the neighboring similar patches in cartoon regions exhibit different spatial configurations from those in texture regions. 

In intuition, the unidirectional distribution of neighboring similar patches around cartoon contours may be attributed to the fact that the often-seen cartoon contours contain a large portion of straight edges. The patches along a straight edge have strong similarity to each other, and the patches on different sides of the edge are often dissimilar. In contrast, the isotropic distribution of neighboring similar texture patches is due to the spatial homogeneity of textures, \eg~cobwebbing in fabrics, scales on snakes, and furs of lions. 
For convenience, the prior that neighboring similar texture patches scatter around, is referred to as the isotropic self-recurrence prior for texture components. 

Building the isotropic patch recurrence prior into the Morphological Component Analysis (MCA) framework~\cite{starck2005morphological}, we develop an effective method for cartoon-texture decomposition in both the noiseless and noisy settings.
The basic idea of MCA is to decompose an image into different components such that each component has a sparse representation under a well-designed system. In this paper, by transferring the isotropic patch recurrence from the image domain to the wavelet domain, we construct a nonlocal wavelet system for modeling the texture component of an image. Combining the constructed nonlocal system  for texture with the local wavelet system  for cartoon, we form an augmented system that is discriminative to separate texture and cartoon.
Benefiting from the use of isotropic patch recurrence prior, the proposed approach shows noticeable performance improvement over the state-of-the-art ones in the experiments.


The contribution of this paper is two-fold. First, we propose the isotropy prior of patch recurrence to model the texture component of an image. The traditional patch recurrence prior designed for image recovery only exploits the repeating property of image patches, without any consideration on the discrimination between the cartoon and texture components. In contrast, the proposed isotropy prior additionally considers the spatial distribution of recurrent patches, \ie, how the image patches repeat over space. As a result, it can well distinguish the texture component from the cartoon component.

Second, based on the isotropy prior, an effective approach to cartoon-texture decomposition is proposed. The proposed approach has several advantages over the existing ones. Compared to the low rank prior of texture patches used in~\cite{schaeffer2013low,ono2014cartoon,fan2017cartoon}, the proposed isotropy prior of patch recurrence is more powerful to model textures. Compared with the patch-recurrence-based approaches~\cite{ma2016group}, the proposed one has additional considerations on the spatial configuration of patch recurrence. By subdividing the patch recurrence into the isotropic and anisotropic versions for texture and cartoon respectively, the proposed approach achieves improvement on distinguishing cartoon contour edges from textures. In comparison to the sparse-coding-based approaches~\cite{papyan2017convolutional,gu2017joint} that only use local sparsification systems, the proposed one utilizes both local and nonlocal systems, which can exploit similar patches for the sparse representation of textures. 

The rest of this paper is organized as follows. Section~\ref{sec:related_work} gives a review of the related works. Section~\ref{sec:pre} presents the preliminaries of our work. Section~\ref{sec:our_method} is devoted to the description of the proposed method. The experimental evaluation is presented in Section~\ref{sec:exp}. Section~\ref{sec:conclusion} concludes the paper.

\section{RELATED WORK}\label{sec:related_work}
There have been many approaches proposed for image  cartoon-texture decomposition. The main differences of these approaches lie the mechanisms of modeling cartoon regions and texture patterns.
Regarding the characterization of cartoon parts, a majority of approaches use the total variation (TV) penalties (\eg~\cite{meyer2001oscillating,vese2003modeling,osher2003image,bertalmio2003simultaneous,aujol2005image,yin2005image,yin2005total,gilles2010properties,duval2010mathematical,zhang2016convolutional}), since TV penalties can efficiently encourage piece-wise constant functions with bounded variations (BV). See also a generalized version of the TV penalty in~\cite{han2017cartoon} for processing color fabric images, which takes both the
first and the second derivatives into account for better protecting soft changing of lightness in local regions. Another approaches to cartoon modeling impose the sparse representation on cartoon parts under some system, such as wavelet frames~\cite{starck2005image,fadili2010image} and curvelet systems~\cite{starck2005morphological,maurel2011locally,ono2012image}, which are very suitable for depicting the smooth structures in cartoon components. The relation between these two types of approaches is discussed in~\cite{cai2012image}, which states that variational methods involving the TV penalty can be implemented by the $\ell_1$-norm analysis approaches using spline wavelets. In addition, compared to the multi-scale extension of TV-based approach in~\cite{tadmor2004multiscale}, the wavelet systems naturally allow multi-resolution analysis on cartoon structures.

In comparison to the cartoon component, the texture component is much more challenging to characterize due to its high complexities and large variations. In the past, many studies modeled texture by the patterns with strong oscillations, and they focused on exploiting the spaces of oscillatory functions which are discriminative to the BV space, as well as exploiting the norms that are antagonistic to the TV norm.
A seminal work can be traced back to Meyer's proposal~\cite{meyer2001oscillating}, whose numerical implementation is first given in~\cite{vese2003modeling}. Following this line of research, many studies have been conducted on addressing numerical issues; see \eg~\cite{osher2003image,starck2003image,aujol2005image,aujol2005dual,aujol2006structure,gilles2010properties,duval2010mathematical,gilles2010properties}. There are also other studies developing new models such as linear filtering models~\cite{buades2010fast,buades2016directional} and multi-scale model~\cite{gilles2012multiscale}, or focusing on applications~\cite{bertalmio2003simultaneous,hu2011multiple,ng2013coupled}.
Since noises also exhibit oscillatory patterns, the aforementioned approaches cannot well handle noisy images. Furthermore, as pointed out in~\cite{schaeffer2013low}, these approaches may fail in discovering the small-magnitude but well-patterned textures.

A promising alternative is the sparse-representation-based approach (\eg~\cite{starck2005image,fadili2010image,maurel2011locally,ono2012image,cao2014segmentation,zhang2016convolutional,papyan2017convolutional,gu2017joint}), which assumes textures can be sparsified under certain systems, \ie, a texture patch can be represented by few of atoms in a certain dictionary (referred to as the synthesis model) or sparsified under some transform (referred to as the analysis model).
With a suitable system, either analytic (\eg~local Fourier frame~\cite{maurel2011locally}) or learned (\eg~\cite{zhang2016convolutional,papyan2017convolutional}), the texture component can be effectively extracted by promoting the sparsity of the representation coefficients. In~\cite{gu2017joint}, the analysis model and synthesis model are combined for further improvement.
Since cartoon features such as edges and contours may also have sparse representations under some proper dictionaries, the sparse-representation-based approaches do not work well when the sparsification system for cartoon has high coherence with that for texture. In other words, one key in the sparse-coding-based approaches is the design of effective sparsification systems whose ambiguity between cartoon and texture can be controlled. Interested readers can refer to~\cite{fadili2010image} for a comprehensive overview. Our proposed method in this paper falls into this kind of approaches.

Another line of research on texture modeling is built upon the low rank prior of texture patches. An early inspiring attempt is given by Schaeffer and Osher~\cite{schaeffer2013low}, which assumes the 
texture patches are almost linearly dependent after certain alignment and then applies the low-rank regularization with nuclear norm to the aligned texture patches. In~\cite{fan2017cartoon}, the $logdet$ function is used to replace the nuclear norm in~\cite{schaeffer2013low} for better approximation to the rank function.
Though capable of processing regular textures with global homogeneity, these methods are not suitable for extracting the textures with distortion or with spatially-varying patterns, which are rather typical in real scenes. As a result, such methods may produce undesirable patterned artifacts on natural images. To overcome this problem, Ono and Miyata~\cite{ono2014cartoon} adapted the low rank prior from the global image to a local window, and the low-rank regularization is applied to the patches of texture component within a local block. Moreover, multiple shear operators are employed for better alignment. Nevertheless, this approach may suffer from the under-sampling problem, \ie, the local block may not contain enough similar patterned patches for stable low-rank approximation, especially for the patch locating near the boundary of a texture region. Besides, the used shear operations may involve additional computational cost.

Instead of sampling patches within a local block, Ma~\etal~\cite{ma2016group} proposed to group similar patches and apply low-rank regularization to each patch group for extracting textures. The cartoon component is extracted by enforcing structured sparsity on each group. It is noted that, though with low-rank approximation, this method employs a different prior from the low rank prior in~\cite{schaeffer2013low,ono2014cartoon,fan2017cartoon}. In fact, the prior in~\cite{ma2016group} can be viewed as the nonlocal patch recurrence prior which has been successfully applied to many image restoration tasks; see~\eg~\cite{dabov2006image,ji2010robust,quan2015data}. However, the direct use of the nonlocal patch recurrence prior in cartoon-texture decomposition is not optimal, as the prior may be applicable to both the cartoon component and texture component.
In this paper, we aim at further exploiting the nonlocal patch recurrence for cartoon-texture decomposition.

\section{PRELIMINARIES}\label{sec:pre}
Throughout this paper, bold upper letters are used for matrices, bold lower letters for column vectors, light lower letters for scalars, hollow letters for sets, and calligraphic letters for operators. The notations $\matx{0}$ and $\matx{I}$ denote the zero matrix and the identity matrix with appropriate sizes respectively, and $\textrm{diag}(\vect{x})$ denotes the square diagonal matrix with vector $\vect{x}$ as its main diagonal.
The $\ell_p$ norm of a vector is denoted by  $\|\cdot\|_p$. Given a sequence $\{\vect{y}^{(t)}\}_{t \in\set{N}}$, $\vect{y}^{(t_{0})}$ denote the $t_{0}$-th element in the sequence. For a vector $\vect{x} \in \set{R}^N$ and a set $\set{S} \subset \{1,\cdots,N\}$, $\vect{x}(i)$ denotes the $i$-th element in $\vect{x}$, and $\vect{x}_{\set{S}}$ denotes the vector concatenating all $\vect{x}(i)$ where $i\in \set{S}$. For a matrix $\matx{X}$, $\matx{X}(i,j)$ denotes the element of $\matx{X}$ at the $i$-th row and $j$-th column.

We briefly present some basics of wavelet tight frames in image processing, as they will be used in the proposed method. Interested readers are referred to \cite{shen2010wavelet} for more details.
A wavelet tight frame is generated by the shifts and dilations of a finite set of generators, which can be used to reveal the local structures of images at different scales. The transform using a wavelet tight frame can be implemented by convolutions with
a set of filters that has certain properties. Given a filter $\vect{a}$, define the corresponding convolution matrix $\opt{S}_{\vect{a}}\in \set{R}^{N\times N}$ by
\begin{equation}
[\opt{S}_{\vect{a}}\vect{v}](n)=[\vect{a}*\vect{v}](n), \forall \vect{v}\in \set{R}^N,
\end{equation}
where $*$ denotes the convolution operation.
For a set of wavelet filters $\{\vect{a}_i\}^m_{i=1}$, the analysis operator of the corresponding wavelet tight frame, or called  wavelet transform, is defined by the following matrix form:
\begin{equation}\label{eq:wavelet_trans}
[\opt{S}_{\vect{a}_1(-\cdot)}^\top, \opt{S}_{\vect{a}_2(-\cdot)}^\top, \dots ,\opt{S}_{\vect{a}_m(-\cdot)}^\top]^\top \in \set{R}^{mN\times N}, 
\end{equation}
and the synthesis operator, or called inverse wavelet transform, is defined by the transpose of the analysis operator.

In this paper, we use the single-level linear spline wavelet tight frame, which has the following three filters in 1D case:
$$
\vect{a}_1=\frac{1}{4}(1,2,1)^\top; \ \vect{a}_2=\frac{\sqrt2}{4}(1,0,-1)^\top; \ \vect{a}_3=\frac{1}{4}(-1,2,-1)^\top.
$$
In the 2D case, the wavelet filters can be constructed via the tensor product of the 1D versions. Note that though wavelet tight frames allow multi-scale analysis, in practice we found that the multi-scale analysis brings little benefits to the proposed method.

\section{OUR METHOD}\label{sec:our_method}
\subsection{Overview}
In this paper, we propose an image cartoon-texture decomposition method using Morphology Component Analysis (MCA)~\cite{starck2005morphological}.
Given an arbitrary signal containing multiple layers as a linear combination, MCA separates the layers such that each layer has a good $\ell_1$-sparse representation under some system.
Given a gray-scale image $\vect{f} \in \set{R}^{N}$ composed of the cartoon component $\vect{u} \in \set{R}^{N}$ and texture component $\vect{v} \in \set{R}^{N}$, we
solve the following minimization problem
\begin{equation}\label{eq:model_proposed_noiseless}
\begin{split}
&\min_{\vect{u},\vect{v}} \ \lVert{\textrm{diag}(\vect{\lambda}_1)\matx{W}\vect{u}}\rVert_1 + \lVert{\textrm{diag}(\vect{\lambda}_2)\matx{J}\vect{v}}\rVert_1,
\\
&\text{s.t.} \ \ \vect{u}+\vect{v}=\vect{f},
\end{split}
\end{equation}
where $\matx{W}\in\set{R}^{M\times N}$ and $\matx{J}\in\set{R}^{P\times N}$ are two systems that sparsify the cartoon component and texture component respectively, and $\vect{\lambda}_1 \in \set{R}^M,\vect{\lambda}_2 \in \set{R}^P$ are two regularization parameter vectors.
The above model can be rewritten into a unconstrained form for adapting
the case where image noises are presented. For Gaussian white noise, the minimization problem becomes
\begin{equation}\label{eq:model_proposed_noisy}
\begin{split}
\min_{\vect{u},\vect{v}} \
\lVert{\textrm{diag}(\vect{{\lambda}}_1)\matx{W}\vect{u}}\rVert_1 + &\lVert{\textrm{diag}(\vect{{\lambda}}_2)\matx{J}\vect{v}}\rVert_1 \\
& + \lVert{\vect{f}-(\vect{u}+\vect{v})}\rVert_2^2.
\end{split}
\end{equation}
Note that $\vect{\lambda}_1,\vect{\lambda}_2$ in~\eqref{eq:model_proposed_noisy} are different from those of~\eqref{eq:model_proposed_noiseless}. We use the same notation for them for convenience.
Combining $\matx{W}$ and $\matx{J}$ into an augmented system, both \eqref{eq:model_proposed_noiseless} and \eqref{eq:model_proposed_noisy} can be rewritten into the standard $\ell_1$ norm minimizations with efficient numerical solvers, which is shown in Sec.~\ref{subsect:num_alg}.

The performance of cartoon-texture decomposition using \eqref{eq:model_proposed_noiseless} or \eqref{eq:model_proposed_noisy} is dependent on the accuracy of the systems $\matx{W}$ and $\matx{J}$ in characterizing cartoon and texture. In our scheme, we define $\vect{W}$ by  some wavelet transform, which is very effective in sparsifying piece-wise smooth signals. Since textures often contain dense small edges whose wavelet representation is not sparse, the regularization term $\lVert{\matx{W}\vect{u}}\rVert_1$ can well distinguish cartoon from texture. Note that since low-pass filter does not sparsify images, we omit the low-pass filter in applying $\matx{W}$ by setting the corresponding parts in $\vect{\lambda}_1$ to zeros.
Regarding the definition of $\matx{J}$, we aim at utilizing the patch recurrence of texture to construct an effective system for sparsifying the texture component, which is discussed in the next part.

\subsection{Construction of Sparsification System for Texture}
We construct the system $\matx{J}\in \set{R}^{mN\times N}$ for sparsifying texture components with the following form:
\begin{equation}
\matx{J} = \matx{L}\matx{T},
\end{equation}
where $\matx{T}\in\set{R}^{mN\times N}$ denotes a wavelet frame of the form~\eqref{eq:wavelet_trans}, and $\matx{L}\in\set{R}^{mN\times mN}$ denotes a discrete Laplacian.
When applying $\matx{J}$ to a texture image $\vect{v}\in \set{R}^N$, $\matx{T}$ serves as a local system to obtain a more effective representation from $\vect{f}$, and $\matx{L}$ serves as a nonlocal difference system to generate a sparse representation from $\matx{T}\vect{v}$.
Let $\vect{c}=\matx{T}\vect{v} \in \set{R}^{mN}$.
Though $\vect{c}$ is not sparse which is a distinct property of texture to cartoon, $\matx{L}\vect{c}=\matx{J}\vect{v}$ is sparse due to the nonlocal difference by $\matx{L}$. In fact, the system $\matx{J}$ can be viewed as a nonlocal wavelet frame~\cite{quan2015data} for sparsifying textures.

Next, we turn to the construction of the nonlocal system $\matx{L}$, which is based on the patch recurrence of texture images. 
In the wavelet representation $\vect{c}=[c_1,\cdots,c_Q]$, each element ${c}_i$ is a wavelet coefficient which is calculated with a wavelet filter of small localized support. In other words, $c_i$ corresponds to a local texture pattern within a patch, which is denoted by $\vect{p}_i$.\footnote{Note that $\vect{p}_i$ is not the $i$th patch in image. Instead, it is the image patch involved in the calculation of the coefficient $\vect{c}_i$ in convolution.}
As a result, the wavelet coefficients calculated on similar texture patches should be similar, implying that the patch recurrence prior can be transferred from the spatial domain to the wavelet domain.
With such a wavelet recurrence prior, for each coefficient ${c}_i$, we can find a set of coefficients $\vect{c}_{\set{S}_i}$ such that
\begin{itemize}
	\item $\set{S}_i \ \subset \ \{1,2,\cdots,Q\} \backslash {i}$;
	\item $\vect{c}_i$ and $\vect{c}_j$ are two wavelet coefficients generated by the same wavelet filter;
	\item  $\vect{p}_j$ is similar to $\vect{p}_i$ for all $j$ in $\set{S}_i$.
\end{itemize}
In other words, $\vect{c}_{\set{S}_i}$ contains the wavelet coefficients whose corresponding texture patches are similar to that of $c_i$. For convenience, we denote the texture patches associated to $\set{S}_i$ by
$
\set{P}_i = \{\vect{p}_j:j\in\set{S}_i\}.
$

To utilize the patch recurrence prior, we first define
$\matx{L}$ by a discrete normalized Laplacian as follows:
\begin{equation}\label{eq:non local operator}
\matx{L}(i,j)=\frac{1}{\omega (\vect{p}_i,\set{P}_i)}
\left\{
\begin{array}{ll}
\omega(\vect{p}_i,\set{P}_i), & {i=j;} \\
-\omega (\vect{p}_i,\vect{p}_j), &   {j \in\set{S}_i;} \\
0, & \textrm{otherwise,}
\end{array} \right.
\end{equation}
where $0<\omega(\vect{p}_i,\vect{p}_j)\leq 1$ is some similarity measurement on two texture patches $\vect{p}_i$ and $\vect{p}_j$, with the principle $\omega(\vect{p}_i,\vect{p}_i)=1$, and $\omega (\vect{p}_i,\set{P}_i)=\sum_{\vect{q}\in \set{P}_i} \omega (\vect{p}_i,\vect{q})$.
A popular choice of $\omega(\cdot,\cdot)$ in existing patch-recurrence-based (or nonlocal) image recovery approaches  (\eg~\cite{Dong2014Compressive,Tao2017Mixed}) is
\begin{equation}\label{eq:omega}
\omega(\vect{p},\vect{q})=e^{-\lVert \vect{p}-\vect{q} \rVert_2^2 / h},
\end{equation}
where $h$ is a scalar.
The matrix $\matx{L}$ can be rewritten as $\matx{L}=\matx{I}-\matx{E}$ where $\matx{E}(i,j)=\omega (\vect{p}_i,\vect{p}_j) /\omega(\vect{p}_i,\set{P}_i) $ if ${j \in\set{S}_i}$ and $0$ otherwise. With this form, it is easy to see that 
\begin{equation}\label{eq:Lap}
(\matx{L}\vect{c})(i) = c_{i} - \frac{1}{\omega(\vect{p}_i,\set{P}_i)} \sum_{{j}\in \set{S}_i} \omega (\vect{p}_i,\vect{p}_j) c_j.
\end{equation}
In other words, $\matx{L}$ relates the nonlocal similar coefficients.
In the case of perfect patch recurrence where $\vect{p}_j=\vect{p}_i$ for all $j \in \set{S}_i$ and all $i$, by using  $\omega(\vect{p}_i,\vect{p}_j)=1$ and $c_i=c_j$  for all $j \in \set{S}_i$, it is straightforward to show that $(\matx{L}\vect{c})(i)=0$ for all $i$, implying that $\matx{L}$ can perfectly sparsify $\vect{c}$. In the case where the patch recurrence  of texture is imperfect, it is reasonable to assume $(\matx{L}\vect{c})(i)\approx 0$ for most $i$, and thus $\matx{L}$ can serve as an effective sparsification system for $\vect{c}$.

The above construction of $\matx{J}$ is dependent on the construction of $\set{S}_i$ which is to collect similar patches for each $\vect{p}_i$. A simple scheme is searching the spatial neighborhood of $\vect{p}_i$ and using the top-$K$ similar image patches to define $\set{S}_i$. This scheme has been widely used in existing nonlocal approaches, \eg~\cite{dabov2006image,quan2015data}) for image recovery. However, it does not work in our approach, as the patch recurrence prior is also applicable to most cartoon contours, \eg, similar patches can be found along the straight edges of contours. As a result, the sparsification system $\matx{J}$ constructed in this setting may also well sparsify the cartoon component $\vect{u}$, which implies ambiguity in the cartoon-texture decomposition.

In order to remedy such an ambiguity problem, we introduce the isotropic patch recurrence prior on texture, as well as its counterpart (\ie~unidirectional patch recurrence prior on cartoon), to construct $\set{S}_i$. These two priors have been illustrated in Fig.~\ref{fig:prior_isotropy_anisotropy}, stating that though the patch recurrence property exists in both the cartoon and texture components, the directions of recurrences, \ie \ the spatial distributions of similar patches, are much different between the two components. Similar patches on cartoon contours are usually distributed along one direction, while similar patches in texture regions tend to scatter in multiple directions.

\begin{figure}[!hbtp]
	\centering
	\includegraphics[angle=0, width=\linewidth]{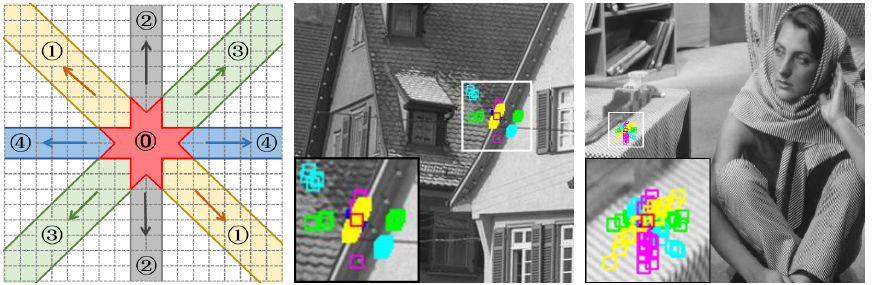}
	\caption{Patch matching using neighborhood partition with $4$ directions. Left: Schematic diagram of the partition strategy. Middle: Matching results on cartoon. Right: Matching results on texture.}
	\label{fig:partition}
\end{figure}

Based on the isotropic recurrence of texture patches, we propose to encode the spatial distribution of similar patterns in defining $\set{S}_i$ to improve the capability of $\matx{L}$ and $\matx{J}$ in distinguishing cartoon elements from texture patterns.
The key idea is to explicitly include similar patches from multiple directions into $\set{P}_i$. With this purpose, we partition the neighborhood $\set{N}(\vect{p})$ of a patch $\vect{p}$ into $D+1$ regions, which includes $D$ banded regions $\{\set{N}^{(j)}\}_{j=1}^D$ along different directions and one central region $\set{N}^{(0)}$ in center. In Fig.~\ref{fig:partition}, we show the partition with $D=4$. For arbitrary $D$, the partition scheme is as follows. Given a patch $\vect{p}$ centered at $\vect{c}$, its neighborhood $\set{N}(\vect{p})$ is defined as a $S\times S$ window $\set{M}$ centered at $\vect{c}$. In $\set{N}(\vect{p})$, we find $D$ banded regions $\{\set{M}^{(j)} \subset \set{N}(\vect{p})\}_{j=1}^D$ which go through $\set{M}$ with center at $\vect{c}$ and are aligned at $0^\circ, \frac{180^\circ}{D}, \cdots, (D-1)\frac{180^\circ}{D}$ respectively. Note that  $\set{M}_1,\cdots, \set{M}_D$ have overlap which is denoted by $\set{M}^{(0)} = \bigcap_{j=1}^D\set{M}^{(j)}$. Then, the partitions $\{\set{N}_j\}_{j=0}^D$ are defined as follow:
\begin{equation}
\set{N}^{(j)}=
\left\{
\begin{array}{ll}
\set{M}^{(0)}, & j=0,\\
{\set{M}^{(j)}}\cap{\tilde{\set{M}}^{(0)}},& j=1,\cdots,D.
\end{array}\right.
\end{equation}

 

With the partition, we search the top-$K$ similar patches of $\vect{p}_i$ in each $\set{N}^{(j)}(\vect{p}_i)$, and the indexes of the similar patches are collected as $\set{S}_i^{(j)}$. See Fig.~\ref{fig:partition} for an example of matched patches by  the proposed partition matching scheme. Then, instead of using $\cup_{j=0}^D \set{S}_i^{(j)}$ to construct the Laplacian operator $\matx{L}$, which is often adopted in traditional patch-recurrence-based approaches, we define a Laplacian operator $\matx{L}^{(i)}$ on each $\set{S}_i^{(j)}$ via~\eqref{eq:Lap}, and the new Laplacian operator $\hat{\matx{L}}$ is defined as the linear combination of $\matx{L}^{(i)}$ for all $i$:
\begin{equation}\label{eq:Lap_new}
\hat{\matx{L}} = \frac{\matx{L}^{(0)} + \matx{L}^{(1)} + \cdots + \matx{L}^{(D)}}{D+1},
\end{equation}
which is a combined Laplacian of the ones with directional grouping. Accordingly, the system $\matx{J}$ is modified to be
\begin{equation}\label{eq:texture_system}
\matx{J}=\hat{\matx{L}}\matx{T}.
\end{equation}

Compared to the traditional Laplacian operator $\matx{L}$ in~\eqref{eq:Lap} that directly uses the top-$K$ patches similar patches in neighborhood to calculate the difference,
the new Laplacian operator $\hat{\matx{L}}$ enforces to use the top-$K$ similar patches from each direction for the calculation. This can be verified by calculating each element of $\hat{\matx{L}}$, which is as follow:
\begin{equation}\label{eq:Lap_new_ij}
\hat{\matx{L}}(i,j)=\frac{1}{D+1}
\left\{
\begin{array}{cl}
D+1, & i=j;\\
-\frac{\omega (\vect{p}_i,\vect{p}_j)}{\omega (\vect{p}_i,\set{P}_i^{(0)})}, & \textrm{$j\in \set{S}_i^{(0)}$;} \\
\vdots \\
-\frac{\omega (\vect{p}_i,\vect{p}_j)}{\omega (\vect{p}_i,\set{P}_i^{(D)})}, & \textrm{$j\in \set{S}_i^{(D)}$;} \\
0, & \textrm{otherwise}.
\end{array} \right.
\end{equation}
 Note that there are many types of graph Laplacians defined for various applications~(\eg~\cite{adler2015linear,dong2016learning,yankelevsky2016dual,pang2017graph}). To the best of our knowledge, this is the first time we use directional grouping to define  graph Laplacians for modeling textures.  It can be further seen that, since each $\matx{L}^{(i)}$ is normalized and weighted with the same factor in the summation, the patch recurrence along each direction has equal contribution to the construction of $\hat{\matx{L}}$. This construction scheme is very useful for distinguishing texture from cartoon,
Consider a cartoon patch $\vect{p}$ on a straight edge whose similar patches lie along the direction of the edge. 
The original version of $\matx{J}$ can generate sparse coefficients from $\vect{p}$ because the operator $\matx{L}$ only uses the similar patches along the edge to represent $\vect{p}$. In contrast, sparse results can be avoided by using the new version of $\matx{J}$, as the operator $\hat{\matx{L}}$ enforces the use of patches along different directions which may be dissimilar to  $\vect{p}$. When $\vect{p}$ is the patch in the texture part with similar patches scattered in the neighborhood, the new version of $\matx{J}$ can inherit the sparsification capability from its old version, because  each $\matx{L}^{(i)}$ in the definition of $\matx{L}$ can utilize $\vect{p}$'s similar patches. 

\subsection{Numerical Algorithm}\label{subsect:num_alg}
Following the MCA framework, we define an augmented system $\matx{D}=[\matx{W} \ \matx{0}; \ \matx{0} \ \matx{J}]$ and then rewrite the minimization problem
~\eqref{eq:model_proposed_noisy} into
\begin{align}
& \min_{\vect{x}}  \ \lVert \textrm{diag}(\vect{\lambda}) \matx{D}\vect{x} \rVert_1, \quad \text{s.t.} \ \matx{A}\vect{x}=\vect{f}, \label{eq:reformulated_noiseless} \\
& \min_{\vect{x}} \ \lVert \matx{A}\vect{x}-\vect{f} \rVert_2^2+\lVert \textrm{diag}({\vect{\lambda}})\matx{D}\vect{x} \rVert_1, \label{eq:reformulated_noisy}
\end{align}
where $\vect{x}=[\vect{u}^\top,\vect{v}^\top]^\top$,
$\matx{A}=[\matx{I} \ \matx{I}]$,
$\vect{\lambda}=[\vect{\lambda}_1^\top,\vect{\lambda}_2^\top]^\top$.

There are many numerical schemes that can be adopted to solve the minimization problems in~\eqref{eq:reformulated_noiseless} and~\eqref{eq:reformulated_noisy}, \eg~ADMM.
In this paper, we use the split-Bregman iteration method~\cite{goldstein2009split}, which can be viewed as another form of ADMM, to develop the numerical solvers due to its simplicity in the form and in the parameter setting.
Regarding the unconstrained problem~\eqref{eq:reformulated_noisy}, we first rewrite it as the constrained problem
\begin{equation}\label{eq:split_noisy}
\min_{\vect{x}} \ \lVert \matx{A}\vect{x}-\vect{f} \rVert_2^2 + \lVert \textrm{diag}(\vect{\lambda}) \vect{d} \rVert_1,
\quad
\st \vect{d}=\matx{D}\vect{x},
\end{equation}
which is the ``splitting" step to separate $\vect{x}$ from the $\ell_1$ penalty. Next, the Bregman iteration is used to solve~\eqref{eq:split_noisy} as follows:
\begin{equation}\label{eq:bregman_noisy}
\left\{
\begin{array}{l}
\displaystyle
(\vect{x}^{(k+1)},\vect{d}^{(k+1)}) =\argmin_{\vect{x},\vect{d}} \ \lVert \matx{A}\vect{x}-\vect{f} \rVert_2^2 + \\
\quad \quad \quad \lVert \textrm{diag}(\vect{\lambda}) \vect{d} \rVert_1 + \gamma
\lVert \matx{D} \vect{x}-\vect{d}+\vect{b}^{(k)} \rVert_2^2 \\[5pt]
\displaystyle
\vect{b}^{(k+1)} =\vect{b}^{(k)}+\delta(\matx{D}\vect{x}^{(k+1)}-\vect{d}^{(k+1)})
\end{array}
\right.
\end{equation}
for $k=0,1,\cdots$, where $\gamma>0$ and $0<\delta\le1$ are two parameters arising from the Bregman iteration. By decomposing the problem in~\eqref{eq:bregman_noisy} into two subproblems, we further rewrite~\eqref{eq:bregman_noisy} as follows:
\begin{equation}\label{eq:bregman_noisy_2}
\left\{
\begin{array}{l}
\displaystyle
\vect{x}^{(k+1)} =\argmin_{\vect{x}} \ \lVert \matx{A}\vect{x}-\vect{f} \rVert_2^2 + \gamma
\lVert \matx{D}\vect{x}-\vect{d}^{(k)}+\vect{b}^{(k)} \rVert_2^2 \\[3pt]
\displaystyle
\vect{d}^{(k+1)} =\argmin_{\vect{d}} \ \lVert \textrm{diag}(\vect{\lambda})\vect{d}  \rVert_1 + \gamma
\lVert \matx{D} \vect{x}^{(k+1)}-\vect{d}+\vect{b}^{(k)} \rVert_2^2 \\[3pt]
\displaystyle
\vect{b}^{(k+1)} =\vect{b}^{(k)}+\delta(\matx{D}\vect{x}^{(k+1)}-\vect{d}^{(k+1)})
\end{array}
\right.
\end{equation}
In~\eqref{eq:bregman_noisy_2}, the subproblem regarding $\vect{x}$ is a least square problem with the analytic solution given by
\begin{equation}\label{eq:least_square}
\begin{split}
\vect{x}^{(k+1)}=(\matx{A^{\top} A} + & \gamma \matx{D^{\top}D})^{-1} \\
(\matx{A^{\top}}\vect{f} & +\gamma\matx{D^{\top}}(\vect{d}^{(k)}-\vect{b}^{(k)})),
\end{split}
\end{equation}
which is calculated by the conjugate gradient method in our implementation to avoid computing the inverse. The subproblem regarding $\vect{d}$ is separable to each dimension of $\vect{d}$ and thus has the analytic solution given by
\begin{equation}\label{eq:thresholding}
\vect{d}^{(k+1)}=T_{\vect{\lambda}}(\matx{D}\vect{x}^{(k+1)}+\vect{b}^{(k)}),
\end{equation}
where $T_{\vect{\lambda}}(\cdot)$ is the soft-thresholding operation defined by
\begin{equation}
(T_{\vect{\lambda}}(\vect{x}))(k)=sign(\vect{x}(k))max(\lvert \vect{x}(k) \rvert-\lambda_k,0).
\end{equation}
Combining \eqref{eq:bregman_noisy_2}, \eqref{eq:least_square} and \eqref{eq:thresholding}, the problem~\eqref{eq:reformulated_noisy} is solved by the following iteration:
\begin{equation*}\label{eq:noisy iteration}
\left\{
\begin{array}{l}
\vect{x}^{(k+1)}=(\matx{A^{\top} A}+ \gamma \matx{D^{\top}D})^{-1}
(\matx{A^{\top}}\vect{f}+\gamma\matx{D^{\top}}(\vect{d}^{(k)}-\vect{b}^{(k)}))\\
\vect{d}^{(k+1)}=T_{\vect{\lambda}}(\matx{D}\vect{x}^{(k+1)}+\vect{b}^{(k)})\\
\vect{b}^{(k+1)}=\vect{b}^{(k)}+\delta(\matx{D}\vect{x}^{(k+1)}-\vect{d}^{(k+1)})
\end{array}
\right.
\end{equation*}
The overall algorithm is summarized in Algorithm~\ref{alg:outline}.

Regarding the noiseless model~\eqref{eq:reformulated_noiseless} which has an additional constraint compared to~\eqref{eq:reformulated_noisy}, we can solve it with a similar scheme which applies an additional Bregman iteration to handle the fidelity constraint.

\begin{algorithm}
	\caption{\label{alg:outline}Numerical scheme for solving~\eqref{eq:reformulated_noisy}}
	\textbf{Input}: Image $\vect{f}$ \\
	\textbf{Output}: Cartoon component $\vect{u}$, Texture component $\vect{v}$\\
	\textbf{Main procedure}:
	\begin{enumerate}
		\item Construct the system $\matx{D}=[\matx{W} \ \matx{0}; \  \matx{0} \ \matx{J}]$;
		\item $\vect{d}^{(0)}:=\matx{D}\vect{x}^{(0)},  \vect{b}^{(0)}=\vect{e}^{(0)}:=\vect{0}$;
		\item For $k=0,\cdots,K-1$,
		\begin{enumerate}

			\item 
				$
				\left\{
				\begin{array}{l}
				\vect{x}^{(k+1)}=(\matx{A^{\top} A}+ \gamma \matx{D^{\top}D})^{-1}\\
				\qquad \qquad (\matx{A^{\top}}\vect{f}+\gamma\matx{D^{\top}}(\vect{d}^{(k)}-\vect{b}^{(k)}))\\
				\vect{d}^{(k+1)}=T_{\vect{\lambda}}(\matx{D}\vect{x}^{(k+1)}+\vect{b}^{(k)})\\
				\vect{b}^{(k+1)}=\vect{b}^{(k)}+\delta(\matx{D}\vect{x}^{(k+1)}-\vect{d}^{(k+1)})
				\end{array}
				\right.
				$
		\end{enumerate}
		\vskip 8pt
		\item $\vect{u}=[\matx{I}, \matx{0}]\ \vect{x}^{(K)}$ 
		and $\vect{v}=[\matx{0}, \matx{I}]\ \vect{x}^{(K)}$.
	\end{enumerate}
\end{algorithm}

\subsection{Adaption of Regularization Parameters}
\begin{figure}[!hbtp]
	\centering
	\newcommand{\IMGWIDTH}{0.87\linewidth}
	\includegraphics[width=\IMGWIDTH]{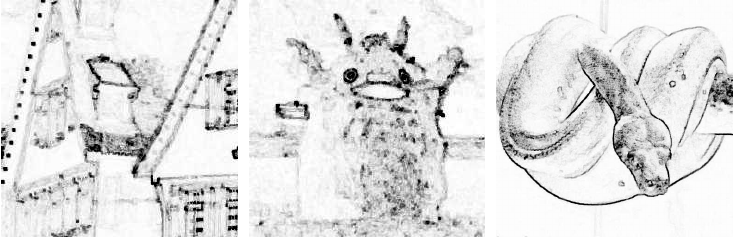}
	\caption{Illustration of the regularization parameter vector $\vect{\lambda}_1$ on the images 'House' (left),  'JackStraw' (middle) and 'Snake' (right). For better visibility, the contrast is adjusted.}
\label{fig:reg_par}
\end{figure}

The regularization parameters $\vect{\lambda}_1$ and $\vect{\lambda}_2$ are crucial to the performance of the proposed models.
Ideally, the values of $\vect{\lambda}_1$ should be high at texture regions and low at contour regions. In analogy, $\vect{\lambda}_2$ should have low values at texture regions and high values at contour regions. 
To effectively estimate the regularization parameters, we propose the following scheme:
\begin{equation}\label{eq:reg_par_sel}
\left\{
\begin{array}{ll}
\vect{\lambda}_1(i)=\beta_1 e^{-\eta_1\phi(i)}, \\
\vect{\lambda}_2(i)=\beta_2 (1-e^{-\eta_2\phi(i)}),
\end{array} \right.
\end{equation}
where $\beta_1,\beta_2$ are the weights of the regularization terms, $\eta_1$ and $\eta_2$ are two scalars for adjusting the shape of exponential decay, and $\phi(i)$ is the function to measure the {"texturelessness"} which is defined by
\begin{equation}\label{eq:texturelessness}
\phi(i)=\sum_{j\in \set{G}_i}|(\matx{J}\matx{W}\vect{u})(j)|^2,
\end{equation}
where $\set{G}_i=\{j:\vect{x}_i$, $\vect{x}_j$ correspond to the same spatial location of the image but different wavelet filters\}. In implementation, since the truth of $\vect{u}$ is unavailable, we use $\vect{u}^{(k)}$ as the estimate of $\vect{u}$.
Note that we initialize $\vect{u}^{(0)}=\vect{f}$ and $\vect{v}^{(0)}=\vect{0}$. Such an initialization scheme is often used in many existing methods (\eg{\cite{ono2014cartoon},\cite{papyan2017convolutional}}).
The exponential function in~\eqref{eq:texturelessness} is to map the $\phi$ to $(0,1]$. See Fig.~\ref{fig:reg_par} for an illustration of $\vect{\lambda}_1$. It is noted  that since low frequency coefficients of cartoon part are not sparse, the corresponding values in $\vect{\lambda}_1$ are set to $\vect{0}$.

\section{EXPERIMENTS}\label{sec:exp}
\subsection{Protocols and Settings}
The performance of the proposed method in cartoon-texture decomposition is evaluated with both the noiseless and noisy settings.
In the noiseless setting, the input images are clean with an ignorable amount of noise, and we solve the decomposition problem~\eqref{eq:model_proposed_noiseless}. In comparison, the noisy setting involves noisy images for testing the performance of~\eqref{eq:model_proposed_noisy}, which also can be viewed as the application to image denoising. When ground-truth images available, we use PSNR and SSIM to measure the  the visual quality of the decomposition results.


Several representative or state-of-the-art approaches are selected for comparison, including
\begin{itemize}
	\item ADMGB (Alternating Direction Method with Gaussian Back Substitution)~\cite{ng2013coupled}, where the TV norm and its dual are used to characterize cartoon and texture respectively;
	\item BNN (Block-wise Nuclear Norm)~\cite{ono2014cartoon}, a representative method utilizing the low-rank property of texture patches;
	\item GID3D (Group-based Image Decomposition 3D)~\cite{ma2016group}, which groups similar image patches and conducts low-rank decomposition on each patch group;
	\item SCSC (Slice-based Convolutional Sparse Coding)~\cite{papyan2017convolutional}, which uses convolutional sparse dictionary learning to discover the underlying patterns of cartoons and textures.
	\item JCAS (Joint Convolutional Analysis and Synthesis)~\cite{gu2017joint}, which integrates the analysis operator and synthesis operator in convolutional sparse coding to better model cartoons and textures.
\end{itemize}
\newcommand{\resultList}[1]{
	#1{ADMGB}\hfill
	#1{BNN}\hfill
	#1{GID3D}\hfill
	#1{SCSC}\hfill
	#1{JCAS}\hfill
	#1{Baseline}\hfill
	#1{Ours}
}
Note that ADMGB is the very recent development on using the dual norm of TV norm for cartoon-texture decomposition. BNN and GID3D are two effective methods based on low-rank approximation. Compared to BNN, GID3D exploits the groups of similar patches, which is closely related to our work. Besides, SCSC and JCAS are also the closely-related works, as they are two most-recent sparse-representation-based methods. 
To test whether the proposed isotropy prior of patch recurrence prior is useful, we construct a baseline method (denoted by 'Baseline') which solves~\eqref{eq:model_proposed_noiseless} without using the isotropy prior, $\ie$, it simply uses the nonlocal system $\matx{L}$ defined in~\eqref{eq:non local operator}. The parameters of Baseline are finely tuned for fair comparison.

In our implementation, the wavelet systems $\matx{W}$ and $\matx{T}$ are set to be the linear spline wavelet system with single level. In the construction of the Laplacian $\hat{\matx{L}}$, the size of search window $S$ is set to $51$, the number of directions $D$ is set to 4 with 16 matched patches in each direction, and the bandwidth parameter $h$ in~\eqref{eq:omega} is set to $0.3$. For the pure decomposition, the parameters $(\beta_1,\beta_2)$ in adaptive weighting are set to $(0.30,0.36)$, and the parameters $(\eta_1,\eta_2)$ in \eqref{eq:reg_par_sel} are both set to 0.5. In the split-Bregman iteration, the parameter $\gamma$ is set to $0.1$, the update step $\delta$ is  set to $1$, and the maximal iteration number $K$ is set to $15$. For noisy image decomposition, the parameters are set as follows: $(\beta_1,\beta_2)=(10^{-5},10^2)$, $(\eta_1$,$\eta_2)=(0.5,0.5)$, $\gamma=2.5$, $\delta=1$, $K=20$.
To improve the patch matching results on noisy images,
a pre-denoising is applied before matching. For fair comparison, we use the same pre-denoising scheme as~\cite{ma2016group}.

\subsection{Decomposition on Noiseless Images}
\label{exp-noiseless}
\subsubsection{\bf Synthetic data}
\begin{figure*}[!hbtp]
	\centering
	\includegraphics[width=0.84\linewidth]{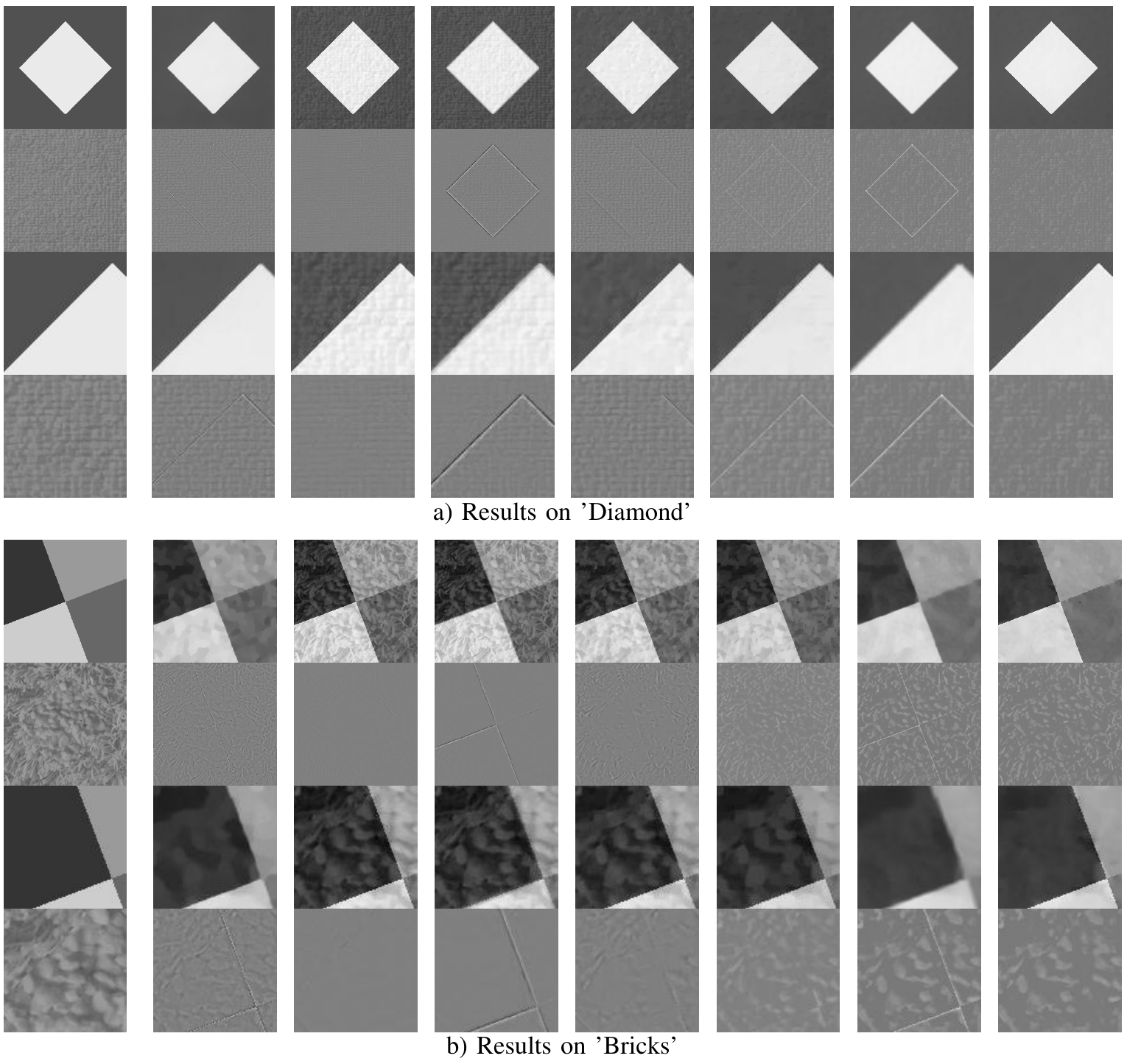}
	\caption{\label{fig:result_diamond_bricks}Decomposition results on 'Diamond' and 'Bricks' (\ie~the $1$st and $14$th synthetic images). The columns from left to right correspond to the ground-truth, the results produced by ADMGB, BNN, GID3D, SCSC, JCAS, Baseline and the proposed method. The rows from top to bottom correspond to the cartoon components, texture components and their close ups.}
\end{figure*}

To evaluate the basic performance of our method, we constructed a set of $16$ synthetic images. Due to space limitations, the details and most results on these images are given in our supplementary materials.
In Fig.~\ref{fig:result_diamond_bricks}(a), we show the decomposition results on synthetic 'Diamond' by different methods.
The benefit of using our isotropy prior is demonstrated by our superior results over Baseline. In the close-ups of the cartoon components, the edges of the diamond are broken and blurred by Baseline, while well preserved by our method. In the close-ups of the texture components, the contour edges of the diamond  appear clearly in the result of Baseline, while correctly rejected by our method.
In the comparison with other methods, our method still show noticeable improvement. Regarding the cartoon components, GID3D and JCAS blur edges, while BNN, GID3D and SCSC preserve some texture patterns. Regarding the texture components, ADMGB, GID3D and SCSC produce visible cartoon edges, while BNN and JCAS extract the textures incorrectly. Another example is shown in Fig.~\ref{fig:result_diamond_bricks}(b), where our method performs the best, generating a clear cartoon component with sharp edges and a clear texture component without visible cartoon edges.

To quantify the performance of different methods, we show in Table~\ref{tb:syn} the average PSNR values and average SSIM on the decomposition results,
where our method performs the best in both terms of PSNR and SSIM.
In summary, both the visual inspection and quantitative evaluation have demonstrated the effectiveness of our isotropic patch recurrence prior.

\begin{table}
	\caption{\label{tb:syn} Average PSNR (dB) and SSIM values of the decomposition results on synthetic images by different methods.}
	\renewcommand{\arraystretch}{1.1}
	\centering
	\newcommand{\PreserveBackslash}[1]{\let\temp=\\#1\let\\=\temp}
	\newcolumntype{L}{>{\PreserveBackslash\centering}p{2 cm}}
	\newcolumntype{M}{>{\PreserveBackslash\centering}p{0.8 cm}}
	\newcolumntype{R}{>{\PreserveBackslash\centering}p{1 cm}}
	\begin{tabular}{@{}L@{}@{}R@{}@{}R@{}@{}R@{}@{}R@{}@{}R@{}@{}R@{}@{}R@{}}		
\hline
Criterion       & ADMGB &  BNN	& GID3D &  JCAS	&  SCSC	&Baseline& Ours \\
		\hline
PSNR (Catoon)   & 28.64 & 25.23 & 24.51 & 27.53 & 26.71 & 28.51 & \textbf{32.29} \\
SSIM (Catoon)   & 0.884 & 0.596 & 0.587 & 0.775 & 0.733 & 0.921 & \textbf{0.937} \\
PSNR (Texture)  & 26.09 & 25.23 & 24.49 & 26.71 & 26.76 & 27.01 & \textbf{27.82} \\
SSIM (Texture)  & 0.789 & 0.591 & 0.558 & 0.761 & 0.779 & 0.784 & \textbf{0.801} \\
		\hline		
	\end{tabular}
\end{table}	

\begin{figure}[!hbtp]
	\centering	
	\includegraphics[width=\linewidth]{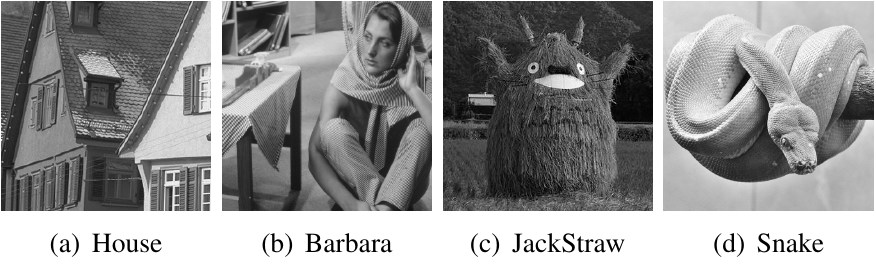}
	\caption{\label{fig:img} Natural images for noiseless decomposition.}
\end{figure}

\subsubsection{\bf Real data}
We use the upper four natural images in Fig.~\ref{fig:img} for performance test, where 'House', 'Barbara', and 'JackStraw' are cited from~\cite{ono2014cartoon,ma2016group,papyan2017convolutional} for fair comparison. These images contain different types of textures (\eg~handmade {\it vs.} natural, regular {\it vs.} random), which can test different aspects of our method. Since PSNR and SSIM are not computable due to the unavailability of ground truths, we compare the results by visual inspection with three criteria. Firstly, contour edges should only appear in the cartoon component. Secondly, well-patterned features should only appear in the texture component. Thirdly, no artifacts are presented in both the components. 

\begin{figure*}[!hbtp]
	\centering
	\newcommand{\PAGEWIDTH}{0.85\linewidth}
	\includegraphics[width=\PAGEWIDTH]{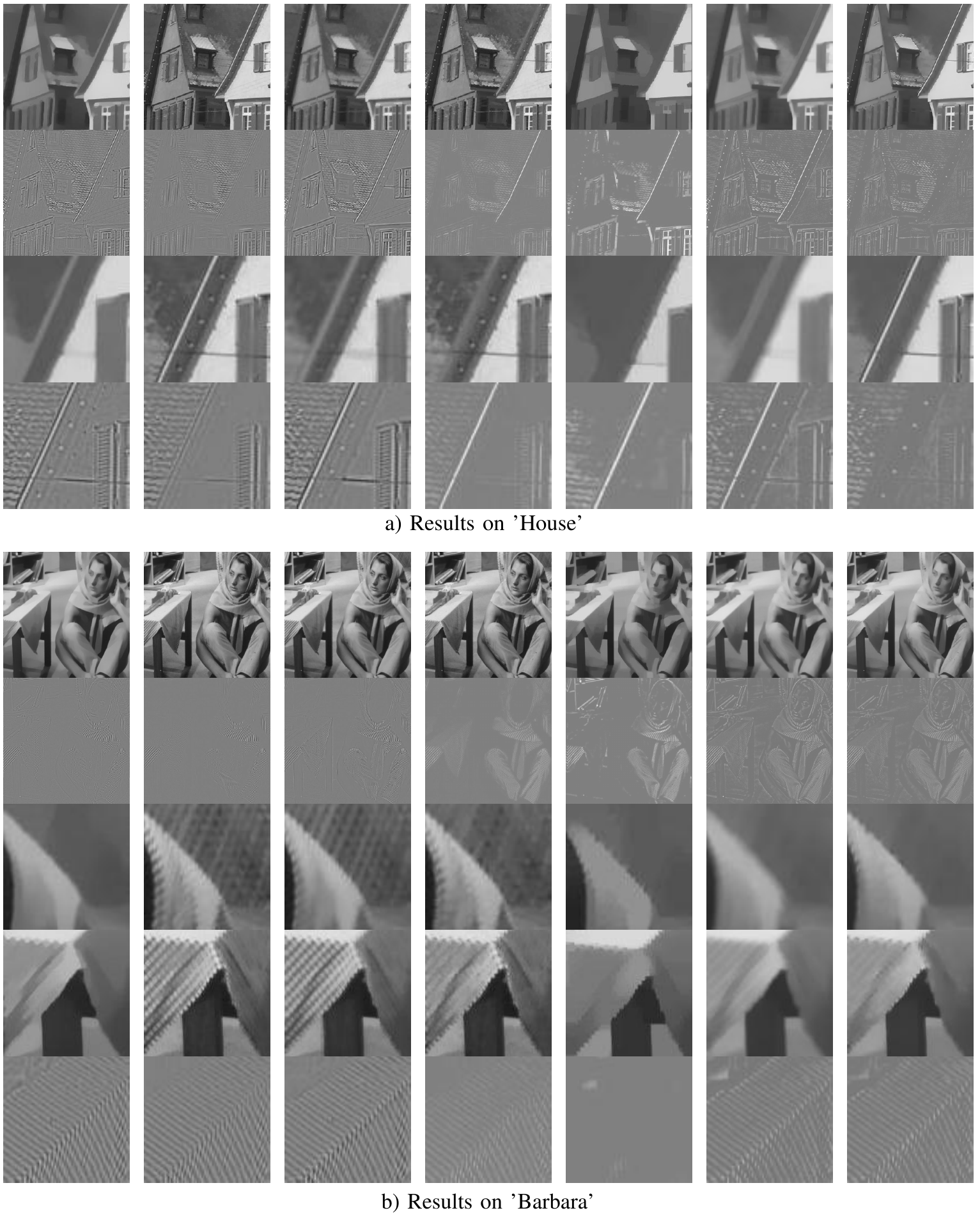}
	\caption{\label{fig:result_house_barbara}Decomposition results on 'House' and 'Barbara'. The columns from left to right correspond to the results by ADMGB, BNN, GID3D, SCSC, JCAS, Baseline and the proposed method respectively. The rows from top to bottom correspond to the cartoon components, texture components and their close ups.}
\end{figure*}

The results on 'House' are shown in Fig.~\ref{fig:result_house_barbara}(a). We can see that our method is good at preserving the contour edges in cartoon component while rejecting them in the texture component. This can be more clearly demonstrated in the close ups of the eave that contains sharp cartoon edges. In our results, the eave is completely kept in the cartoon component. In contrast, the other compared methods except BNN assign a strong edge along the eave to the texture component, resulting in blur or even removal of the eave in cartoon. In particular, the two patch-recurrence-based methods GID3D and Baseline, which do not utilize the isotropy prior, obviously blur the eave.

The results on 'Barbara' are shown in Fig.~\ref{fig:result_house_barbara}(b). We can see that the proposed method is capable of extracting well-patterned features in the texture component and rejecting them in the cartoon component.
To show the advantages of our method over others, we display the close ups of the table cloth and the chair behind the woman in the figure. In the cartoon components of both table cloth and chair, only ADMGB, SCSC, Baseline and our method can completely remove the textures. 
Compared to SCSC which fails to extract the cloth textures, our method correctly handles the textured patterns. Compared with ADMGB and Baseline that yield blurred cartoon components, our method preserves more details in cartoon, which can be seen more clearly by zooming at the face of Barbara.
Notice that our method slightly loses the face details in the cartoon component. The reason is that the face is of low-resolution and contains dense, short and weak edges, for which the isotropic patch recurrence prior is inapplicable.

\begin{figure*}[!hbtp]
	\centering
	\newcommand{\PAGEWIDTH}{0.85\linewidth}
	\includegraphics[width=\PAGEWIDTH]{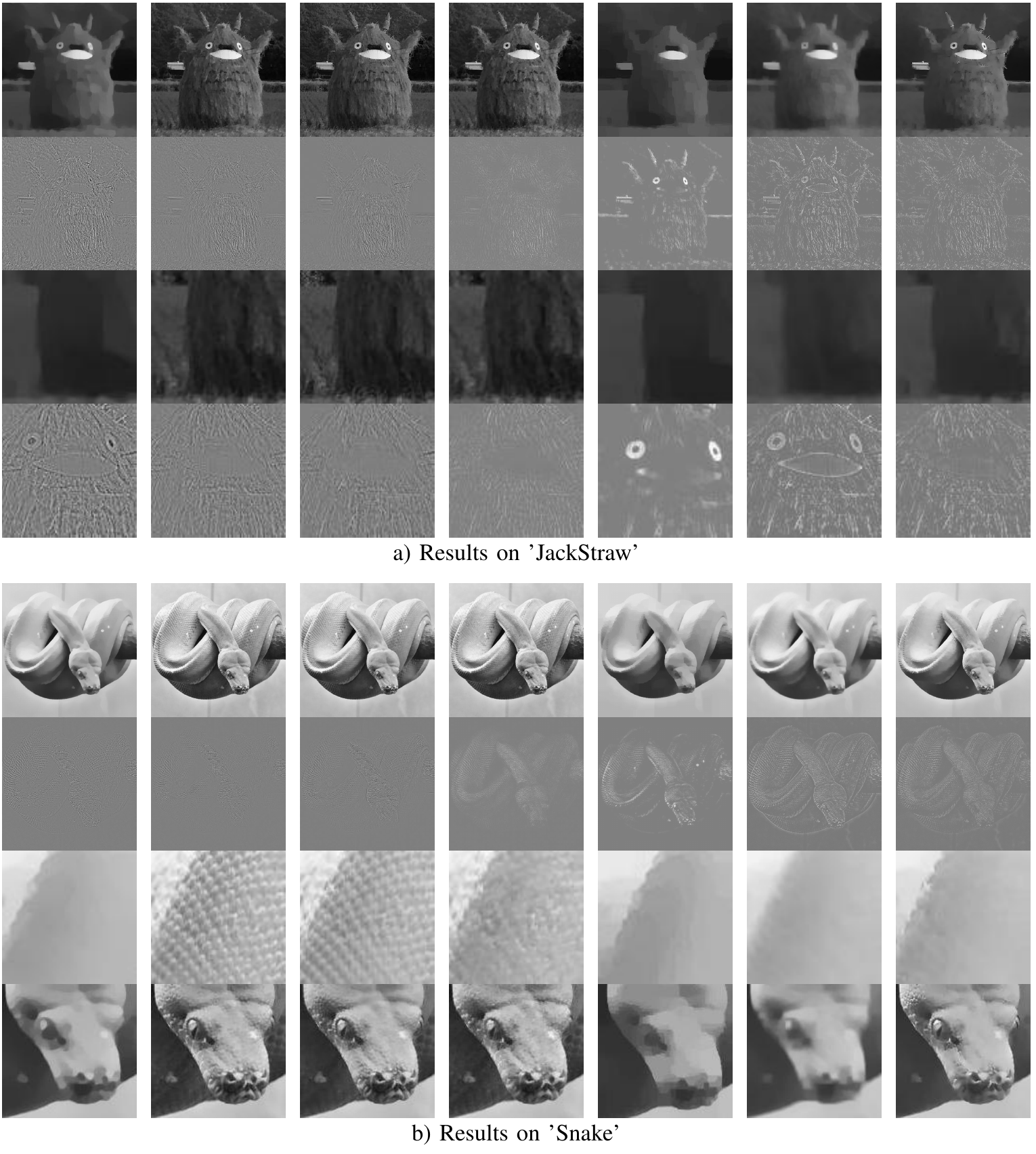}
	\caption{\label{fig:straw_snake}Decomposition results on 'JackStraw' and 'Snake'. The columns from left to right are the results by ADMGB, BNN, GID3D, SCSC, JCAS, Baseline and the proposed method respectively. The rows from top to bottom correspond to the cartoon components, texture components and their close ups.}
\end{figure*}

The results on 'JackStraw' and 'Snake' are shown in Fig.~\ref{fig:straw_snake}. Regarding 'JackStraw', it can be seen that all the compared methods except ours either introduce the straw textures into the cartoon component or assign the contours of mouth and eyes to the texture component.  In comparison, our method can simultaneously handle the cartoon component and texture component well.
Similar phenomena can be found in the results of 'Snake'. In the cartoon components all the compared methods except ours either reject the snake texture correctly but meanwhile delete the head details, or preserve the head details well but meanwhile keep most snake texture. Compared with these methods, the proposed one simultaneously handles the both. 

In summary, all the visual results on natural images have demonstrated the superior performance of our method over the existing ones, as well as the effectiveness of the isotropic patch recurrence prior in cartoon-texture separation.

\begin{figure}[!hbtp]
	\centering
	\includegraphics[width=0.69\linewidth]{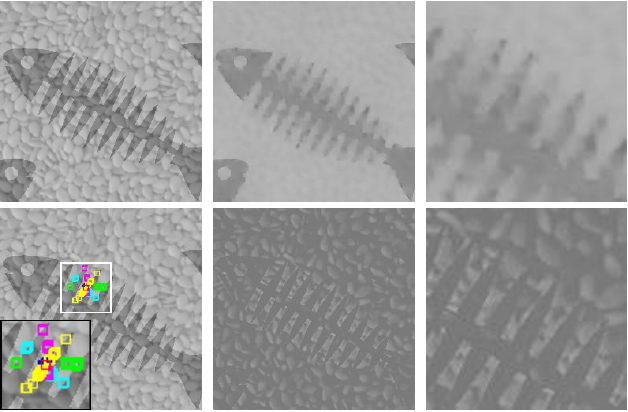}
	\caption{\label{fig:result_fishbone}Decomposition results on 'Fishbone'. The columns in the first row (form left to right, up to down) correspond to the input image, extracted cartoon component, close-up of cartoon component the illustration of patch matching results, extracted texture component, and close-up of texture component.}
\end{figure}

\vspace{12 px}
\noindent{\bf Failure case and limitation.}
We show a failure case of our method in Fig.~\ref{fig:result_fishbone}, where the input image is composed by a cartoon image of a fish bone and a natural texture image. 
It can be seen that our method blurs the contour of the bones in the cartoon component and fails to reject these contours in the texture component. The reason is probably that the patch recurrence on the bone contour tend to be isotropic so that they are misidentified as a texture patch by our isotropy prior of patch recurrence on texture components.


\newcommand{\WIDTH}{1\linewidth}

\subsection{Decomposition on Noisy Images}


To evaluate the performance of our method in noisy settings, we use four noisy images for test, which are shown in the supplementary materials due to space limitations.
Among the test images, "Barbara-1",  "Barbara-2" and "Roof" are cited from~\cite{ono2014cartoon,ma2016group} for fair comparison. 
Following~\cite{ma2016group}, the original images are degraded by adding a Guassian noise with zero mean and standard deviation $\sigma=0.1$ and then used as inputs. As a result, each input image contains three components: cartoon, texture and noise. We separate an input image into these three components by solving~\eqref{eq:model_proposed_noisy}.
For comparison, BNN and GID3D are selected, as they have published results on the test images  with available codes. The quality of the results are evaluated by visual inspection, as well as in terms of PSNR and SSIM.

\begin{figure*}[!hbtp]
	\centering
	\newcommand{\PAGEWIDTH}{0.88\linewidth}
	\includegraphics[width=\PAGEWIDTH]{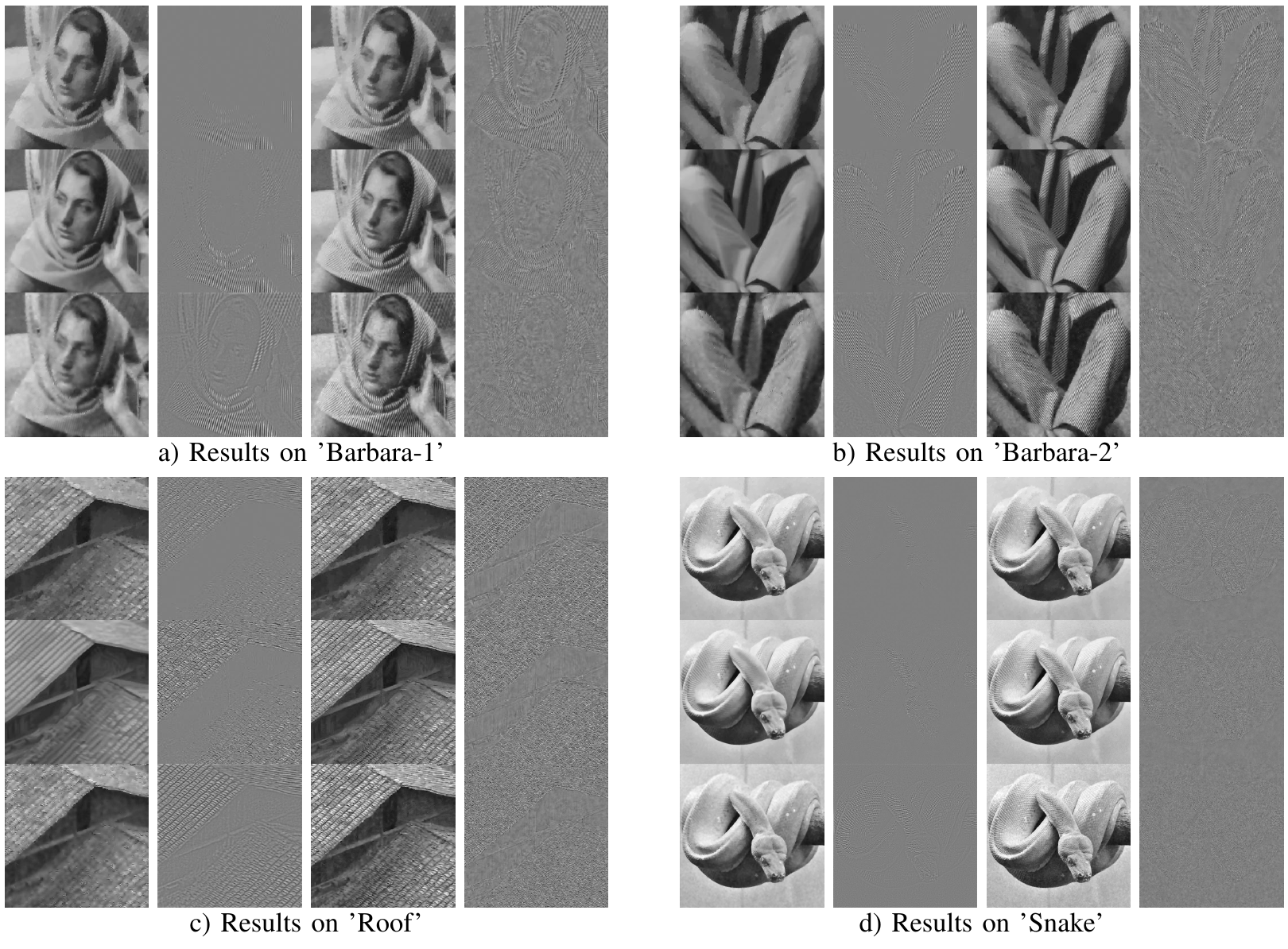}
	\caption{\label{fig:denoise}Decomposition results on noisy images. The rows from top to bottom correspond to BNN, GID3D and the proposed method. The columns from left to right to the cartoon components, texture components, denoised images, and noise components.}
\end{figure*}

The decomposition results are shown in~Fig.~\ref{fig:denoise}. We can see that our method performs well even in the presence of noises, with better visual quality than other methods.
Notice the results of BNN and GID3D. The textures of the chair in Fig.~\ref{fig:denoise}(a) and the textures of the snake in Fig.~\ref{fig:denoise}(d), are presented in cartoon instead of the texture components. In contrast, our method  extracts the texture components correctly in these cases.
Note that our method may assign some unexpected flat patches to the texture components, \eg~the face of Barbara, the wall of the house, and the background behind the snake. The reason is probably that noisy flat patches are non-sparse in the wavelet domain and they may be similar to the neighboring patches, which misleads the judgment of our method.

Fig.~\ref{fig:denoise} also shows the denoised images, which are calculated by removing the noise component from the input. Compared with BNN and GID3D, our method preserves more details but with a few slight artifacts. We also evaluate the results by the randomness of noise components, which is a criterion used in~\cite{buades2005review}. The less structures appear in the noise component (implying higher degree of randomness), the lower coherence exists between noise components and other components, and the better the denoising performance is. It can be seen that, compared with those of other methods, our produced noise components are more random with lower correlation to the input images. 
The PSNR and SSIM values of the denoised images are listed in Table~\ref{tb:denoise}, where our method achieves the highest ones on all the images expect 'Roof'. 

In conclusion, the proposed method have demonstrated its power in cartoon-texture decomposition on noisy images, with superior performance to other compared methods on both decomposition results and denoising results.

\newcommand{\NWIDTH}{1\linewidth}


\begin{table}[htbp]
	\caption{\label{tb:denoise}PSNR (dB) and SSIM values of denoised images by different methods.}
	\renewcommand{\arraystretch}{1.1}
	\addtolength{\columnsep}{-20mm}
	\centering
	\setlength{\tabcolsep}{3.5pt}
	\resizebox{\columnwidth}{!}{
	\begin{tabular}{@{}c|c@{}c|c@{}c|c@{}c|c@{}c}
		\hline
		\multirow{2}[0]{*}{Method} & \multicolumn{2}{c|}{Barabara-1} & \multicolumn{2}{c|}{Barabara-2} & \multicolumn{2}{c|}{Roof} & \multicolumn{2}{c}{Snake} \\
		& \multicolumn{1}{c}{PSNR} & \multicolumn{1}{c|}{SSIM} & \multicolumn{1}{c}{PSNR} & \multicolumn{1}{c|}{SSIM} & \multicolumn{1}{c}{PSNR} & \multicolumn{1}{c|}{SSIM} & \multicolumn{1}{c}{PSNR} & \multicolumn{1}{c}{SSIM} \\
				\hline
		BNN   & 26.72 & 0.765 & 25.43 & 0.778 & 23.32 & 0.740  & 26.04 & 0.704 \\
		GID3D & 27.99 & 0.820 & 26.51 & 0.824 & \textbf{23.92} & 0.779 & 27.16 & \textbf{0.744} \\
		Ours  & \textbf{28.66} & \textbf{0.834} & \textbf{27.25} & \textbf{0.844} & 23.62 & 0.784 & \textbf{27.17} & 0.733 \\
		\hline
	\end{tabular}%
}
	\label{tab:addlabel}%
\end{table}%

\subsection{Analysis on Influence of Parameters}
The selection of the model parameters $\beta_1$ and $\beta_2$ is crucial, as it controls the regularization on each component. To analyze the influence of these parameters, we compared the decomposition results generated by different values $\beta_1$ and $\beta_2$. Due to space limitations, the details and results are given in our supplementary materials. It is observed that increasing $\beta_1$ or decreasing $\beta_2$ will move the elements from texture components moved to cartoon components, and vice versa. We also observe that both the decomposition results and the denoising results are stable to the small changes of $\beta_1$ and $\beta_2$.

The number of level of wavelet system in our method  is also an interesting parameter to test. In the experiments, we found that using multi-scale wavelets brings little benefits on the test images. Considering the additional computational burden from using multiple scales, we use single-scale in this paper. In the supplementary materials, we show some results of the multi-scale extension of our method.

\subsection{Investigation: Decomposition with Missing Pixels}
Since the proposed isotropy prior of patch recurrence works well in the decomposition with denoising, it is interesting to see how it performs in other inverse problems. In the last part of this section, we extend our method to handle image inpainting, \ie~decomposition with missing pixels. The extension is intuitive, and its details are left in the supplementary materials due to space limitations.

\begin{figure}[!hbtp]
	\centering
	\includegraphics[width=0.69\linewidth]{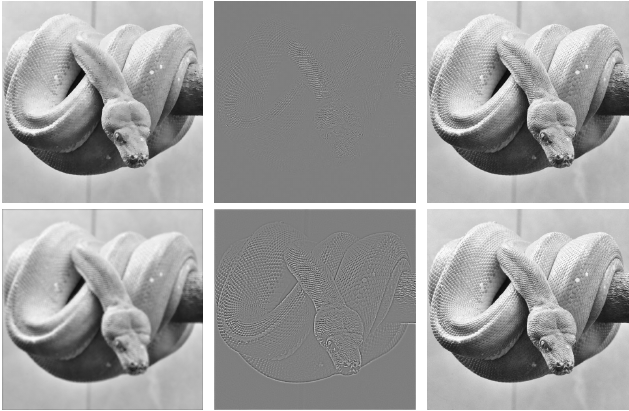}
	\caption{\label{fig:result_snake_inpainting}Decomposition results on 'Snake' with $40\%$ pixels missing by BNN (top) and the proposed method (bottom). The columns from left to right correspond to the cartoon components, texture components and the recovered image.}
\end{figure}

%
Our method is compared with BNN which has published inpainting results with available code. The quality of results are evaluated both visually and quantitatively. Due to space limitations, we only show the visual results of 'Snake' in Fig.~\ref{fig:result_snake_inpainting}. We can see that BNN leaves almost all textures of the snake in cartoon, making its texture component rather weak. In comparison, our method produces much clearer cartoon component and texture component. The results of other images can be referred to the supplementary materials.
The PSNR and SSIM values are listed in Table\ref{tb:inpainting}. Our method outperforms BNN by a large margin on the test images.

\begin{table}
	\caption{\label{tb:inpainting}PSNR (dB) and SSIM values of inpainted images by different methods.}
	\renewcommand{\arraystretch}{1.1}
	\centering
	\begin{tabular}{cccccc}		
		\hline
		Method	& Measure &Barbara-1 & Barbara-2 & House & Snake\\
		\hline
		BNN  & PSNR & 32.46 & 30.24 & 28.98 & 28.87 \\
		BNN  & SSIM & 0.958 & 0.952 & 0.922 & 0.905 \\
		Ours & PSNR & \textbf{35.71} & \textbf{31.88} & \textbf{30.37} & \textbf{30.85} \\
		Ours  & SSIM & \textbf{0.976} & \textbf{0.961} & \textbf{0.944} & \textbf{0.934} \\
		\hline
	\end{tabular}
\end{table}

To sum up, we have shown that our method can be adapted to handling images with missing pixels. Our method may be also extended to deal with other inverse problems. However, since this paper focus on image decomposition, we leave the topic of inverse problem to our future work.

\section{CONCLUSION and FUTURE WORK}\label{sec:conclusion}
In this paper, we proposed the isotropic patch recurrence prior for modeling the texture component of an image. The proposed  prior can be viewed as a refinement of the traditional patch recurrence prior, which can distinguish textures from cartoons by considering the spatial configuration of recurrent patches. Based on the isotropic patch recurrence prior, we developed an effective approach for cartoon-texture decomposition. The performance of the proposed approach was evaluated in both noiseless and noisy setting. The experimental results show the superior performance of the proposed approach to the state-of-the-art ones, which have demonstrated the power of the proposed approach as well as the effectiveness of the isotropic patch recurrence prior.

In future, we would like to further exploit the isotropic patch recurrence in four directions. Firstly, the patch recurrence may exist not only in spatial domain but also across image scales. Thus, we will investigate the cross-scale isotropy prior of patch recurrence to improve the modeling of textures. Secondly, the unidirectional patch recurrence of cartoon has not been explicitly used in our method, and we will study its exploitation in cartoon-texture decomposition.
Thirdly, we have shown the possible application of the proposed prior to image inpainting. To continue, we would like to investigate the adaption of the proposed method to general image inverse problem. Finally, the images processed in this paper are gray-scale. A straightforward extension to handle color images is applying our method to each color channel of the image. However, such a strategy cannot fully exploit the additional cues existing in color channels. Thus, we would like to investigate how to utilize the isotropic patch recurrence prior for the  decomposition on color images.

\section*{ACKNOWLEDGMENT}

The author would like to thank the support by National Natural Science Foundation of China (61602184,61872151,61672241,61528204), Science and Technology Planning Project of Guangdong
Province (2017A030313376), Science and Technology Program of Guangzhou (201707010147), and Fundamental Research Funds for the Central Universities (x2js-D2181690).

\ifCLASSOPTIONcaptionsoff
\newpage
\fi



\bibliographystyle{IEEEtran}
\end{document}